\documentclass[runningheads]{llncs}
\usepackage[T1]{fontenc}
\usepackage{booktabs}   

\usepackage{graphicx}
\usepackage{xcolor}

\usepackage[normalem]{ulem}
\usepackage{multirow}

\begin{document}
\title{Pretrained Deep 2.5D Models for Efficient Predictive Modeling from Retinal OCT}

\author{Taha Emre$^{1*}$ \and Marzieh Oghbaie$^{2*}$ \and Arunava Chakravarty$^1$ \and Antoine Rivail$^{2}$ \and Sophie Riedl$^1$ \and Julia Mai$^1$ \and Hendrik P.N. Scholl$^{6,7}$ \and Sobha Sivaprasad$^3$ \and Daniel Rueckert$^{4,5}$ \and Andrew Lotery$^8$ \and Ursula Schmidt-Erfurth$^1$ \and Hrvoje Bogunovi\'c$^{2}$}

\authorrunning{Emre et al.}
\titlerunning{Pretrained Deep 2.5D Models for Retinal OCT:}

\institute{Dept. of Ophthalmology and Optometry, Medical University of Vienna, Austria \and 
Christian Doppler Lab for Artificial Intelligence in Retina, Dept. of Ophthalmology and Optometry, Medical University of Vienna, Austria\and
NIHR Moorfields Biomedical Research Centre, Moorfields Eye Hospital NHS Foundation Trust, London, United Kingdom \and
BioMedIA, Imperial College London, United Kingdom \and
Institute for AI and Informatics in Medicine, Klinikum rechts der Isar, Technical University Munich, Germany \and
Institute of Molecular and Clinical Ophthalmology Basel, Switzerland \and
Department of Ophthalmology, University of Basel, Basel, Switzerland \and
Clinical and Experimental Sciences, Faculty of Medicine, University of Southampton, United Kingdom\\
\email{\{taha.emre,marzieh.oghbaie,hrvoje.bogunovic\}@meduniwien.ac.at}}

\maketitle              
\def\thefootnote{*}\footnotetext{These authors contributed equally to this work}\def\thefootnote{\arabic{footnote}}
\begin{abstract}
In the field of medical imaging, 3D deep learning models play a crucial role in building powerful predictive models of disease progression. However, the size of these models presents significant challenges, both in terms of computational resources and data requirements. Moreover, achieving high-quality pretraining of 3D models proves to be even more challenging. To address these issues, hybrid 2.5D approaches provide an effective solution for utilizing 3D volumetric data efficiently using 2D models. Combining 2D and 3D techniques offers a promising avenue for optimizing performance while minimizing memory requirements. In this paper, we explore 2.5D architectures based on a combination of convolutional neural networks (CNNs), long short-term memory (LSTM), and Transformers. In addition, leveraging the benefits of recent non-contrastive pretraining approaches in 2D, we enhanced the performance and data efficiency of 2.5D techniques even further. We demonstrate the effectiveness of architectures and associated pretraining on a task of predicting progression to wet age-related macular degeneration (AMD) within a six-month period on two large longitudinal  OCT datasets.
\end{abstract}
\section{Introduction}
3D imaging modalities are routinely employed in clinics for diagnosis, treatment planning and tracking disease progression. Thus, automated deep learning (DL) based methods for the classification of 3D image volumes can play an important role in reducing the time and effort of medical experts. However, the training and design of 3D classification models are challenging as they are computationally expensive, consume large amount of GPU memory during training and require large training datasets to prevent over-fitting. These issues are further exacerbated by the more recent Vision Transformer (ViT) architectures which have been shown to require significantly larger amounts of training data to outperform CNNs. Yet, in the medical domain, there is often a scarcity of labeled training data, especially in the case of 3D imaging modalities. 

The gold-standard 3D imaging modality in ophthalmology is retinal Optical Coherence Tomography (OCT). It is of particular value in the management of patients with Age-Related Macular Degeneration (AMD), the leading cause of blindness in the elderly population. Although asymptomatic in the intermediate stage (\textit{iAMD}), it may progress to a late stage  known as \textit{wet-AMD}, which is characterized by a significant vision loss. Thus, development of an effective personalized prognostic model of AMD using OCT would be of large clinical relevance.   
Given an input OCT scan of an eye in the \textit{iAMD} stage, we aim to develop efficient 3D prognostic models that can predict whether the eye will progress to the \textit{wet-AMD} stage within a clinically relevant time-window of 6 months, modeling the problem as a binary classification task. The lack of well-defined clinical biomarkers indicative of the future risk of progression, large inter-subject variability in the speed of AMD progression and large class imbalance between the progressors (minority class) and non-progressors (majority class) makes it a challenging machine learning task.

Given the above limitations, 2.5D architecture may be an effective approach for building prognostic models from volumetric OCT: it comprises a 2D network applied to each slice of the input volume followed by a second stage to aggregate the feature representations across the slices. 
Compared to 3D models, the 2D ones can be more effectively pretrained on large labeled natural image datasets such as ImageNet or on an unlabeled in-domain dataset of images of the same imaging modality using Self-Supervised Learning (SSL).


In this work, we analyze the impact of different DL architectures and pretraining schemes in the context of developing an effective prognostic classification model using OCT volumes of patients with AMD. 
We first address the problem of limited data availability with an effective in-domain SSL to pretrain 2D CNN weights. We then address the challenge of processing volumetric data by transferring the pretrained 2D CNN weights to a hybrid 2.5D deep learning framework. 
Our evaluation on two large longitudinal datasets  underscores the advantages of such hybrid approach in 3D medical image analysis, and highlights the importance of in-domain pretraining in low data scenarios. 



\subsection{Related Work}

\subsubsection{Deep neural network architectures for 3D predictions}
3D CNNs employ large 3D isotropic convolutions, making them computationally expensive with a significantly increased number of trainable parameters. This makes them prone to over-fitting with limited training data. Moreover, pretrained models weights are more commonly available for 2D CNNs and they cannot be directly used to initialize the 3D networks for fine-tuning. To tackle these limitations, two main directions have been explored: inflating pretrained 2D CNNs into 3D networks or using  Multiple Instance Learning (MIL) in a 2.5D setting. The first approach is based on the Inflated 3D Convnets~\cite{carreira2017quo} which were proposed as a new paradigm for an efficient video processing network using 2D pretrained weights. They achieved this by \textit{inflating} 2D convolutional kernels along the time dimension with a scaling factor.

MIL is an efficient way of processing 3D volumes or videos~\cite{das2020b}. The main idea is to process each 2D component (slice in a 3D volume or frame in a video) of the 3D data individually using a 2D CNN and then pool their output feature embeddings to obtain a single feature representation for the entire 3D data. Average Pooling is commonly employed to integrate the features from each 2D slice/frame in a linear and non-parametric manner. Alternatively, more complex architectures based on LSTMs have also been explored for pooling. Due to their directionality, vanilla LSTMs are better suited for processing videos by modeling the forward flow of time, and using the output from the last time-step as the feature for the entire video. However, a Bidirectional LSTM (BiLSTM) is required to capture the non-directional nature of 3D OCT volumes~\cite{kurmann2019fused}.

Recently, ViTs replaced CNN based models as state-of-the-art. One downside of ViTs is that they require large training datasets and extensive training duration, which are even amplified with 3D modalities. Indeed, the patch-based 3D/video processing ViTs~\cite{arnab2021vivit,fan2021multiscale,tang2022self} process videos through spatio-temporal attention, and 3D volumes with isotropic 3D images of voxels. On the other hand, it has been repeatedly demonstrated that ViTs benefit from the application of convolutional kernels in the earlier blocks~\cite{chen2021visformer,xiao2021early}. Similar to the LSTM based approaches, hybrid ViTs have been successfully applied to video recognition tasks for memory efficiency and speed. In~\cite{Neimark_2021_ICCV}, the video frames are treated as patches and their embeddings are extracted using a ResNet18 which are forwarded to a transformer model.  ViT hybrids focus on efficiency by first processing 2D instances (frames, cross-sectional volume slices), then obtaining a final score from the ViT which is used as a feature aggregator. In medical imaging, CNN + Transformer hybrids are already being used to address the MIL problems such as in whole-slide images~\cite{shao2021transmil} or histopathology images~\cite{li2021dt}. A 3D transformer ViViTs~\cite{arnab2021vivit} were proposed for video analysis. They deploy different strategies to model the interactions between spatial and temporal dimensions at various levels of the model. One of the modes of ViViT denoted as Factorised Self-Attention (FSA), consists in factorizing the attention over the input dimensions. Each transformer block processes both spatial and temporal dimensions simultaneously instead of two separate encoders which makes the network adaptable for 3D volumes.

\subsubsection{Self supervised pretraining}
SSL pretraining aims to generate meaningful data representations without relying on manual labels. Utilizing pretrained network weights obtained through SSL reduces the requirement for labeled data while simultaneously boosting performance in downstream tasks. It is particularly useful when dealing with noisy and highly imbalanced class labels, as it helps prevent overfitting~\cite{balestriero2023cookbook}. Contrastive learning~\cite{chen2020simple} has emerged as one of the most successful SSL approaches. They aim to learn representations that are robust to expected real-world perturbations, while encoding distinctive structures that enable the discrimination of different instances. To achieve invariance against these perturbations, contrastive methods heavily rely on augmentations that create two transformed images, which the models try to bring together while pushing apart the representations of pairs from different instances.

Emre et al.~\cite{emre2022tinc} adapted contrastive augmentations for 2D B-scan characteristics. Additionally, they proposed to use two different scans of a patient from two different visit date as inputs to the contrastive pipeline. The extra temporal information was exploited through a time sensitive non-contrastive similarity loss, termed as \textit{TINC} loss, to induce a difference in similarity between the pairs based on the time difference between them. In the end, they showed that the time sensitive image representations were more useful in longitudinal prediction tasks. In this study, we used \textit{TINC} as the main pretraining method.

\section{Methods: Predictive Model from Retinal OCT Volume}

Predicting the progression to wet-AMD inherently involves a temporal aspect, as patients typically undergo multiple follow-up scans. However, the practicality of capturing temporal information is constrained by factors such as the availability of regularly scanned patients and computational costs. Therefore, we tackled the task as a binary risk classification problem from a given visit.

\begin{figure}[!ht]
    \centering
    \frame{\includegraphics[width=.75\textwidth]{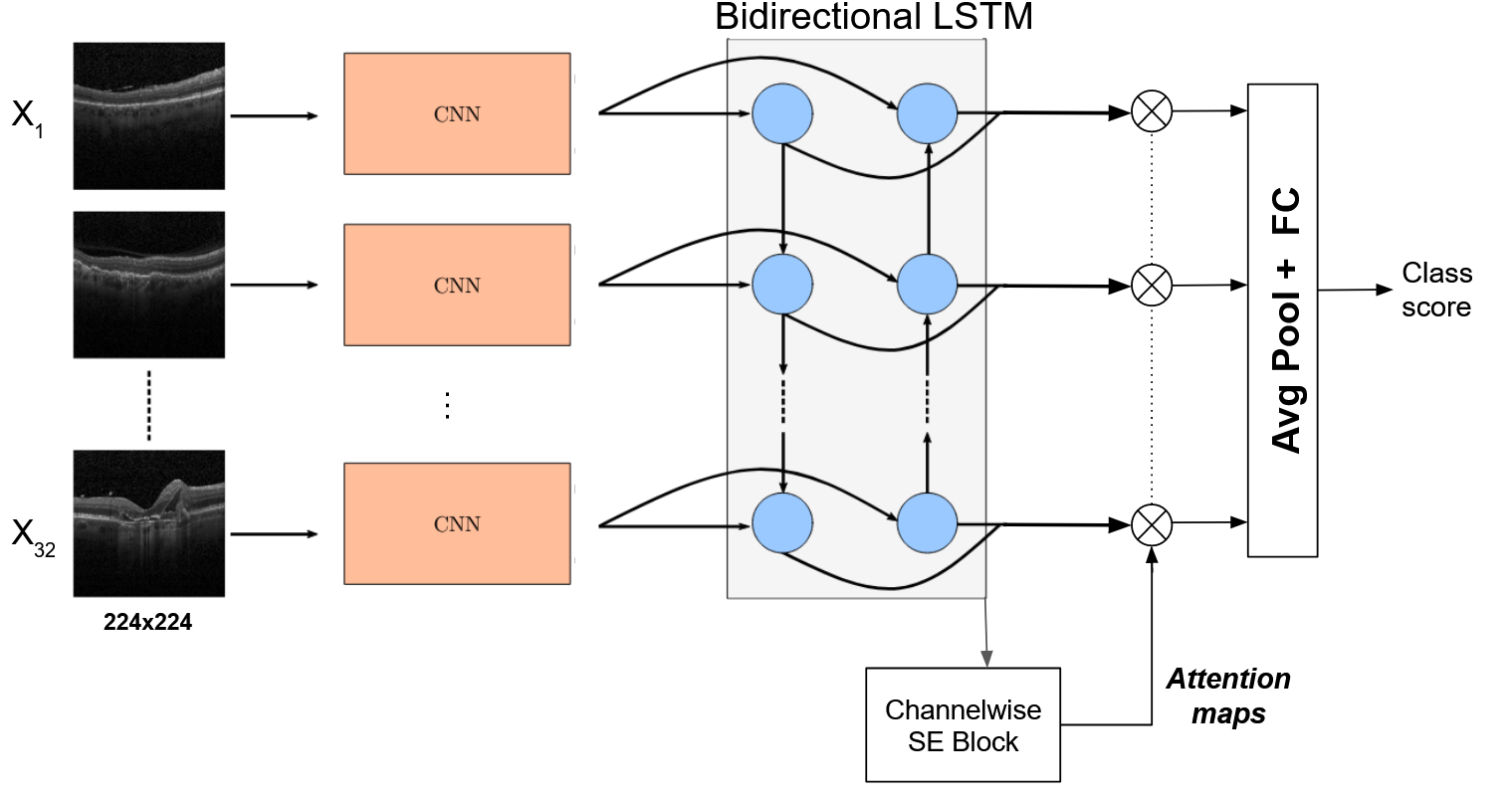}}
    \vspace{0.5cm}
    \frame{\includegraphics[width=.75\textwidth]{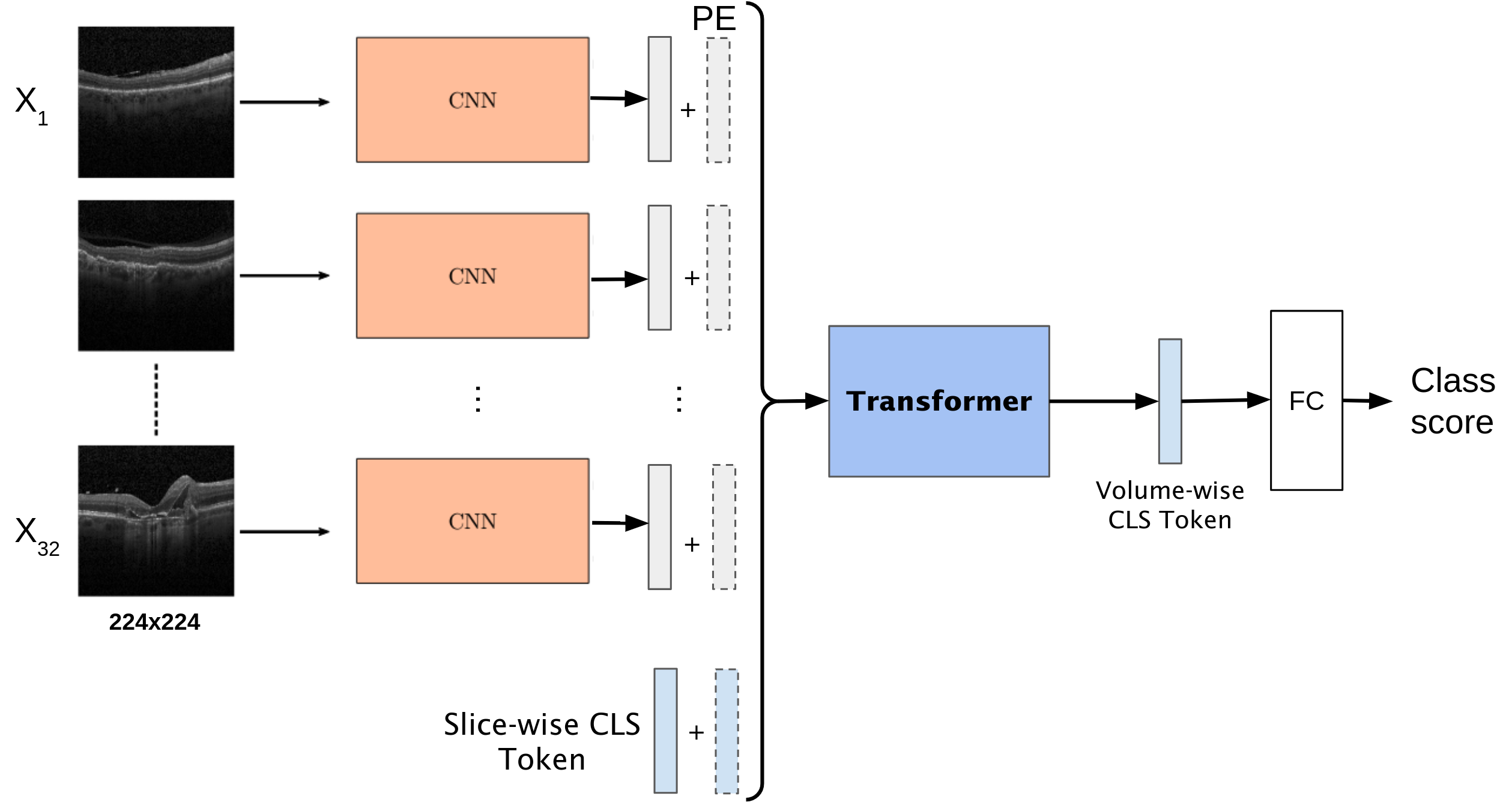}}
    \caption{Architecture of the two evaluated 2.5D approaches: (top) CNN + BiLSTM with attention, (bottom) CNN + Transformer}
    \label{fig:models}
\end{figure}

We explored two hybrid 2.5D architectures (Fig.~\ref{fig:models}), that utilize 2D CNN to generate B-scan embeddings, followed by either an LSTM or a Transformer network to produce volume-level predictions. The CNNs are based on ResNet50 and a B-scan representation is obtained by applying Global Average Pooling on the final feature map of the ResNet50 model, yielding a vector of size 2048. The CNNs were either pretrained in a non-contrastive manner, using \textit{TINC}~\cite{emre2022tinc} on temporal OCT data, or \textit{ImageNet} ResNet50 weights from the torchvision library~\cite{torchvision2016} were used. 


\paragraph{\textbf{CNN + BiLSTM}} Such networks have already been proposed for 3D OCTs~\cite{kurmann2019fused}. Unlike standard LSTMs, BiLSTMs provide two outputs for each ``time step'' (in our case B-scan). Each output can be used for B-scan level prediction, if such labels are available as in~\cite{kurmann2019fused}. However, wet-AMD progression prediction task is a MIL problem where the labels are available per OCT volume. The straightforward choice for aggregating the outputs would be average pooling but we hypothesised that OCT biomarkers indicative of wet-AMD progression are subtle and scarce. Thus, in order to make the predictions more robust, an attention layer attached to the end of BiLSTM. The goal is to enforce B-scan level sparsity with attention such that subtle biomarkers do not blur out due to the global pooling operations. Moreover, we believe the B-scans with the highest attention weights will serve as ideal candidates for clinical inspection to discover novel biomarkers and phenotypes associated with progression to wet-AMD. 

The BiLSTM comprises 32 time-steps (the number of B-scans used from an OCT volume) and a hidden state representation of size 512. For the final attention layer, we used a Squeeze-and-Excitation layer~\cite{hu2018squeeze} adapted for stacked representations (Fig.~\ref{fig:models}). We will refer to this model as CNN + BiLSTM  without explicitly mentioning the attention layer.
\paragraph{\textbf{CNN + Transformer}}
ViTs are known for requiring longer training and larger datasets~\cite{lee2021vision}. Moreover, in-domain pretrained weights for medical images are not as commonly available as they are for natural images. Motivated by this, we developed a hybrid CNN + Transformer model that leverages pretrained weights for the 2D CNN backbone. We attached a small Transformer on top of the 2D B-scan representations to obtain the final prediction. In this approach, each representation is treated as a patch embedding, representing one of the B-scans. Similar to BiLSTM, Transformers can capture the relation between the B-scans. However, unlike BiLSTM, Transformers do not require a pooling operation before the prediction head because they learn a single classification token that encapsulates the necessary information for making the prediction (Fig.~\ref{fig:models}). Additionally, self-attention inherently provides attention scores over the B-scans.

The Transformer network consists of 4 blocks, each with 2 self-attention heads. To handle the large dimension of the patch embedding (2048), the MLP output size in the transformer blocks is reduced to 1024, unlike in standard ViTs that scale up the dimensions in the MLP with respect to the patch embedding dimension. To prevent over-fitting on the small downstream dataset, we inserted Drop Path layers (Stochastic Depth~\cite{touvron2021training}).

\paragraph{\textbf{3D ViViT}}We selected FSA ViViT that follows slow-fusion approach as opposed to hybrid models (late-fusion), where each Transformer block models both spatial and temporal interactions simultaneously. For ViViT, the model is first pretrained for binary OCT volume classification on the public DukeAMD dataset~\cite{Farsiu2014} that includes $269$ intermediate age-related macular degeneration (iAMD) and $115$ normal patients acquired with Bioptigen OCT device.

\paragraph{\textbf{I3D}} When using 2D weights for 3D convolutional kernels, we followed the paradigm of~\cite{carreira2017quo}, and inflated a pretrained ResNet50 
to process OCT volumes.

\section{Experiments}

The scans for the downstream predictive modeling task were manually labeled by ophthalmologists using a clinically relevant time interval of six months. All \textbf{scans} that convert to  \textit{wet}-AMD within the next 6 months have a positive label, while all \textbf{scans} that do not convert within the interval have a negative label. 

\paragraph{Datasets: } The models are trained and evaluated on the fellow-eye dataset from the HARBOR clinical trial\footnote{NCT00891735. https://clinicaltrials.gov/ct2/show/NCT00891735}. It is a longitudinal 3D OCT dataset where each patient is imaged monthly for a duration of 24 months with a Cirrus OCT scanner. Each OCT scan consists of 128, 2D cross-sectional B-scan slices covering a field of view of $6\times6$ mm$^2$. We split the dataset into pretraining and downstream (progression prediction) sets. The pretraining dataset comprises 540 patients and 12,506 scans and also includes images of the late stages of AMD. Among 463 patients, 113 are observed to progress to wet-AMD, yielding only 547 scans with a positive label out of a total of 10,108 scans. The extreme class imbalance makes the progression prediction a very challenging task.

A second longitudinal dataset, PINNACLE~\cite{sutton2022developing}, is used only for the downstream task of progression prediction. Unlike HARBOR, the PINNACLE dataset was acquired with a TOPCON scanner. This resulted in a domain shift due to the difference in the acquisition settings and noise characteristics between the OCT scanner types. With the same labeling criteria, PINNACLE provided 127 converter \textbf{patients} out of 334 (536 positive scans out of 2813). Since the pretraining is only performed on unlabelled scans from the HARBOR dataset, experiments on the PINNACLE set are used to evaluate the performance of the pretraining method 
when the downstream dataset undergoes a domain shift.

In downstream tasks, both datasets were split in the following way: $20\%$ of the patients were kept for hold-out test set, while the remaining scans in each dataset formed the training sets. A stratified 4-fold cross-validation at a patient-level was carried out on the training sets for hyper-parameter tuning. Treating each of the 4 folds as the validation set and the remaining data for training, resulted in an ensemble of 4 models. The mean and standard deviation of the performance of the 4 models on the hold-out test set is reported in Table~\ref{Tbl:Results}.

\paragraph{Image Preprocessing} The curvature of the retina was flattened by shifting each A-scan (image column) in the volume such that the Bruch's membrane (extracted using the method in~\cite{botond2022}) lies along a straight plane. Next, we extracted the central 32 B-scans, which were then resized to $224 \times 224$. During the progression prediction training, the input intensities were min-max scaled, followed by translation, small rotation, and horizontal flip as data augmentations. The same preprocessing was applied to both HARBOR and PINNACLE datasets. For the in-domain SSL-based pretraining, we followed the contrastive transformations outlined in \cite{emre2022tinc}. 

\paragraph{Training Details} CNN+BiLSTM models were fine-tuned using an ADAM optimizer with a batch-size of 20, a learning rate of $0.0001$, which is updated using cosine scheduler, and a weight decay of $10^{-6}$. Similarly, in I3D experiments, we used ADAM with a batch-size of 64, a learning rate of $0.001$, which is updated using cosine scheduler, and a weight decay of $10^{-6}$. In CNN+Transformer, SGD optimizer with momentum was used, as it was found to perform better than ADAM, with a learning rate of $0.001$ which is updated using cosine scheduler, no weight decay was used. We tested 2.5D models with frozen and end-to-end fine-tuning setup and picked the best. For ViViT, we used Adam optimizer with an initial learning rate of $10^{-5}$ which is updated by the cosine scheduler, no weight decay was used. For this experiment, the batch size is set to $8$.

\section{Results}

\setlength{\tabcolsep}{4pt}
\renewcommand{\arraystretch}{1.2}
\begin{table}[!t]
\centering
\caption{Predictive performance of the evaluated models on the internal HARBOR and the external PINNACLE datasets.}\label{Tbl:Results}
\resizebox{1.0\textwidth}{!}{
\begin{tabular}{cccllll}
\toprule
\multirow{2}{*}{\textbf{Model}}&\multirow{2}{*}{\textbf{\#Params}} & \multirow{2}{*}{\textbf{Pretraining}} & \multicolumn{2}{c}{\textbf{HARBOR}} & \multicolumn{2}{c}{\textbf{PINNACLE}} \\
\cmidrule(lr){4-5} \cmidrule(lr){6-7}
& & & \multicolumn{1}{c}{AUROC} & \multicolumn{1}{c}{PRAUC} & \multicolumn{1}{c}{AUROC} & \multicolumn{1}{c}{PRAUC}\\ \midrule
I3D&$46M$&ImageNet & $0.727\pm 0.012$ & $0.134\pm 0.007 $& $0.602\pm 0.036$ & $0.157\pm 0.026$ \\
I3D &$46M$&\textit{TINC}& $0.750\pm 0.037$ & $\mathbf{0.162 \pm 0.034}$ & $0.644\pm 0.039$ & $0.170 \pm 0.023$ \\
CNN+BiLSTM &$34M$&ImageNet &$ 0.742\pm 0.028$ &$0.153 \pm 0.012$&$ 0.622\pm 0.042$ &$0.164 \pm 0.028$\\
CNN+BiLSTM &$34M$&\textit{TINC}& $\mathbf{0.766\pm0.012}$&$\mathbf{0.153\pm0.003}$&$\mathbf{0.646\pm0.019}$&$\mathbf{0.190\pm0.025}$ \\
CNN+Transf. &$108M$&ImageNet&$0.738\pm0.032 $&$0.152 \pm0.035$&$0.617\pm 0.055$&$0.156 \pm 0.031$\\
CNN+Transf. &$108M$&\textit{TINC}&$\mathbf{0.752 \pm 0.022}$&$0.145\pm0.023$&$\mathbf{0.656 \pm 0.027}$&$\mathbf{0.179\pm0.020}$  \\
ViViT$_\mathrm{FSA}$&$34M$&DukeAMD & $0.628\pm0.064$ & $0.098\pm0.023$& $0.566\pm0.054$ & $0.121\pm0.019$\\
\bottomrule
\end{tabular}}
\end{table}

The performances of four distinct architectures, I3D~\cite{carreira2017quo}, ViViT with FSA~\cite{arnab2021vivit}, and the two proposed hybrid 2.5D models, i.e. CNN+BiLSTM and CNN+Transformer, are presented in Table~\ref{Tbl:Results}. 
Each architecture was initialized either with ImageNet or TINC weights, with an exception of ViViT transformer,which was pretrained on DukeAMD dataset.

Firstly, comparing the two initialization strategies confirms the superiority of TINC pretraining in terms of AUROC score in all cases (Table~\ref{Tbl:Results}). This finding highlights the advantage of in-domain pretraining with limited amount of data over pretrained weights coming from a natural image dataset. Although it is probable that both natural and medical images share low-level features, we should emphasize on the drastic effects of the underlying noise characteristics originating from differences in modalities, standard views of the internal organs and tissues, and limited number of expected variances in medical images, on the model performance.
Most of the cases, \textit{TINC} improves PRAUC. On HARBOR dataset, CNN + Transformer with TINC is not better than ImageNet in terms of PRAUC score. This can be due to the fact that PRAUC is more sensitive to differences in probabilities, while AUROC is more concerned with the correct ranking of predictions, which is more relevant for progression prediction. The PINNACLE dataset, characterized by a strong domain shift caused by the intrinsic properties of the scanner, showed that TINC pretraining consistently outperformed ImageNet pretraining, despite a significant drop in AUROC range compared to the HARBOR dataset (Table~\ref{Tbl:Results}).

When we compared the architectures, it is clear that 2.5D approaches outperform both of the 3D models. The CNN + BiLSTM model has significantly fewer trainable parameters than the CNN + Transformer model (34M vs 108M) with a similar number of FLOPs (130G and 133G, respectively). Despite its smaller size, CNN + BiLSTM outperformed CNN + Transformer in Table~\ref{Tbl:Results} for AUROC and PRAUC. In external data experiments on PINNACLE, it achieved comparable results for AUROC while outperforming CNN + Transformer for PRAUC in Table~\ref{Tbl:Results}. This suggests that BiLSTM methods still have merit in the era of Transformers, especially under conditions such as limited and imbalanced 3D data. This can be attributed to the more data requirement of ViTs which affects CNN + Transformer model as well. It is important to highlight that both hybrid models outperformed the I3D model due to their explicit modeling of the relationship between individual B-scans and their better utilization of high-level representations.
Similar to I3D, experiments regarding FSA ViViT also confirm that the simultaneous processing of both dimensions in the input volume did not provide extra benefit over the corresponding counterparts with the same number of parameters ($\sim$34M), indicating the advantage of less complicated models (CNN + BiLSTM/Transformer) for this specific task. 

\section{Conclusion}

In this work, we performed a systematic evaluation of hybrid 2.5D models which utilize already available pretrained 2D backbones. Our results demonstrate that 2.5D approaches not only exhibit efficient memory and label usage, but also outperform larger 3D models when suitably pretrained. The addition of an attention layer to CNN + BiLSTM provides attention scores which in turn can facilitate model explainability. 
Thus, we conclude that deep learning models consisting of 2D CNNs in combination with LSTM continue to offer merits in predictive medical imaging tasks with limited data, outperforming both 2.5D and 3D ViTs. Furthermore, the in-domain pretraining approach \textit{TINC} consistently outperformed the approaches with ImageNet-pretrained weights, highlighting the importance of domain information for predictive tasks. 
These findings provide valuable insights for further development of accurate and efficient predictive models of AMD progression in retinal OCT.

\subsubsection{Acknowledgements} This work was supported in part by Wellcome Trust Collaborative Award (PINNACLE) Ref. 210572/Z/18/Z, Christian Doppler Research Association, and FWF (Austrian Science Fund; grant no. FG 9-N)

\bibliographystyle{splncs04}
\bibliography{ref}
\end{document}